\documentclass{article}

\usepackage{microtype}
\usepackage{graphicx}
\usepackage{subfigure}
\usepackage{booktabs}

\usepackage{hyperref}

\usepackage{tcolorbox}

\usepackage[accepted]{icml2024}

\usepackage{amsmath}
\usepackage{amssymb}
\usepackage{mathtools}
\usepackage{amsthm}
\usepackage{multirow}
\usepackage{enumitem}

\usepackage[capitalize,noabbrev]{cleveref}

\theoremstyle{plain}

\theoremstyle{definition}

\theoremstyle{remark}

\icmltitlerunning{Gradual Divergence for Seamless Adaptation: A Novel Domain Incremental Learning Method}

\begin{document}

\twocolumn[
\icmltitle{Gradual Divergence for Seamless Adaptation: A Novel Domain Incremental Learning Method}

\icmlsetsymbol{equal}{*}

\begin{icmlauthorlist}
\icmlauthor{Kishaan Jeeveswaran}{tue}
\icmlauthor{Elahe Arani}{equal,tue,wayve}
\icmlauthor{Bahram Zonooz}{equal,tue}
\end{icmlauthorlist}

\icmlaffiliation{tue}{Dep. of Mathematics and Computer Science, Eindhoven University of Technology, NL}
\icmlaffiliation{wayve}{Wayve Technologies Ltd, London, UK}

\icmlcorrespondingauthor{}{j.kishaan@tue.nl}

\icmlkeywords{Lifelong Learning, Continual Learning, }

\vskip 0.3in
]



\printAffiliationsAndNotice{\icmlEqualContribution} 

\begin{abstract}
Domain incremental learning (DIL) poses a significant challenge in real-world scenarios, as models need to be sequentially trained on diverse domains over time, all the while avoiding catastrophic forgetting. Mitigating representation drift, which refers to the phenomenon of learned representations undergoing changes as the model adapts to new tasks, can help alleviate catastrophic forgetting. In this study, we propose a novel DIL method named \textit{DARE}, featuring a three-stage training process: \underline{D}ivergence, \underline{A}daptation, and \underline{RE}finement. This process gradually adapts the representations associated with new tasks into the feature space spanned by samples from previous tasks, simultaneously integrating task-specific decision boundaries. Additionally, we introduce a novel strategy for buffer sampling and demonstrate the effectiveness of our proposed method, combined with this sampling strategy, in reducing representation drift within the feature encoder. This contribution effectively alleviates catastrophic forgetting across multiple DIL benchmarks. Furthermore, our approach prevents sudden representation drift at task boundaries, resulting in a well-calibrated DIL model that maintains the performance on previous tasks.
\footnote{\small Code at \url{https://github.com/NeurAI-Lab/DARE}.}
\end{abstract}



\section{Introduction}

Domain incremental learning (DIL) is a subset of continual learning (CL) that addresses the challenge of acquiring knowledge from new domains or tasks in an incremental manner without forgetting previously acquired knowledge. DIL holds significance in real-world applications like autonomous driving and robotics, where data distribution can shift due to factors like changing weather condition and location
~\cite{mirza2022efficient}. Deep neural networks (DNNs) suffer from the problem of catastrophic forgetting, where the weights of the network associated with old tasks are overwritten by new information, resulting in a decline in performance for previously learned tasks.


\begin{figure}[t]
    \centering
    \centerline{\includegraphics[width=0.475\textwidth]{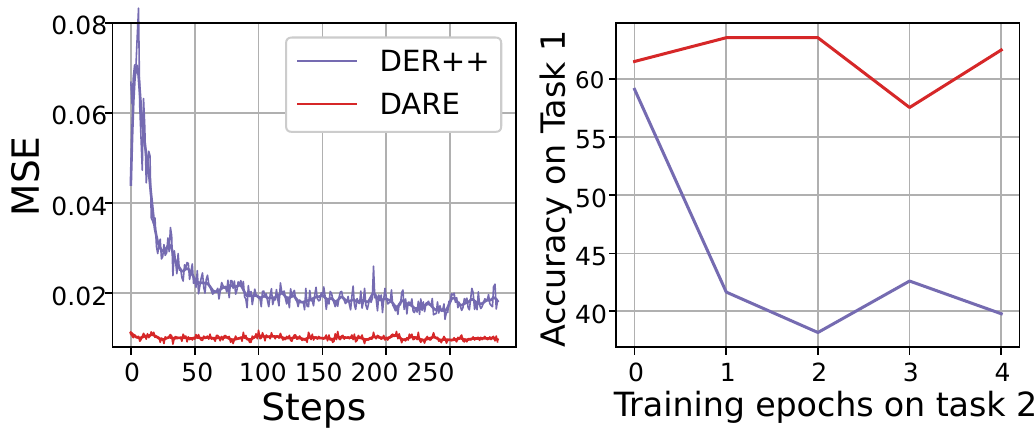}}
    \caption{Relationship between representation drift and task 1 accuracy on DN4IL dataset with buffer size 50. The representations of buffered samples, mainly belonging to the first domain, experience an abrupt drift at the task boundary, which is directly associated with the decrease in accuracy.}
    \label{fig:motivation}
\end{figure}

Various approaches have been proposed to alleviate catastrophic forgetting, which can be grouped into three main categories: interleaving past task samples during new task learning (Experience Replay)~\cite{ratcliff1990connectionist, rebuffi2017icarl}, constraining the change in weights of DNNs pertinent to past tasks (Regularization)~\cite{kirkpatrick2017overcoming, li2017learning}, or expanding the architecture with new branches for learning new tasks without overwriting existing task parameters (Architecture Expansion)~\cite{rusu2016progressive, fernando2017pathnet}. Although experience replay has been shown to effectively mitigate catastrophic forgetting, it does not explicitly address drift in representations at task boundaries caused by the disruption of clustered representations corresponding to previously learned tasks~\cite{caccia2022new}.
Representation drift is directly correlated with performance drop on old tasks and contributes significantly to catastrophic forgetting (see Figure \ref{fig:motivation}).

DNNs aim to acquire clustering representations for similar classes in each task. However, the clusters formed by previous tasks may shift when learning new classes, resulting in a decline in accuracy for old tasks, as observed in new domain learning in DIL (Figure \ref{fig:motivation}). The issue of representation drift is addressed in CL literature through various methods. \citet{caccia2022new} proposed isolating new samples from the old buffer samples and employing distinct loss functions to mitigate drift. This approach hinges on separating old and new classes and the quality of negative samples for the proposed semisupervised loss, limiting its applicability in DIL scenarios. Furthermore, representation drift in DIL remains unexplored in the existing literature. To address this, we suggest a three-stage training process to gradually adapt the learning model to new domain sample representations.


Concretely, we propose a novel approach for mitigating abrupt representation drift and catastrophic forgetting in DIL by adapting the representations of new domain samples into the feature space spanned by the old domains. The proposed method employs a three-stage training process (\textbf{D}ivergence, \textbf{A}daptation, \textbf{RE}finement) while learning new domains. During the Divergence and Adaptation stages, the model clusters the representations of new domain samples into the feature space spanned by the first domain, while the Refinement stage helps the model learn the new domain samples. Our method, DARE, helps to mitigate changes to the representations of old domains, resulting in better overall accuracy. Furthermore, we propose an effective buffer sampling strategy to integrate the proposed algorithm into the CL framework. By using this strategy, we can store important samples in the buffer that capture the maximum information about the ``dark knowledge" between data samples.
Specifically, our contributions are as follows:
\begin{itemize}[noitemsep,topsep=0pt]
    \item We propose a novel domain incremental learning method to effectively adapt the representations of the new domain into the feature space spanned by the prior domains.
    \item We, through extensive analyses, demonstrate the effectiveness of our method in mitigating forgetting, task recency bias, and suppressing detrimental representation drifts at task boundaries in DIL.
    \item We propose and employ an effective buffer sampling strategy that maximizes the information stored in the buffer without significant memory overhead.
\end{itemize}

\section{Related Works}

\textbf{Domain Incremental Learning.}
DIL studies the ability of DNNs to continually adapt to new domain data while preserving performance on prior domains, such as adapting to different weather conditions~\cite{mirza2022efficient}. Many approaches in DIL rely on learning task-specific information and plugging it during inference.
DISC~\cite{mirza2022efficient} stores domain-specific batch norm statistics and uses them during inference to detect objects under different weather conditions. \citet{garg2022multi} use a dynamic architecture for domain incremental segmentation by learning both domain-invariant and domain-specific parameters. However, these approaches require task-id during inference, which violates the CL desiderata~\cite{farquhar2018towards}.

Approaches to address catastrophic forgetting in CL can be broadly divided into three categories: regularization-based~\cite{kirkpatrick2017overcoming, li2017learning}, parameter isolation~\cite{rusu2016progressive, fernando2017pathnet}, and rehearsal-based~\cite{ratcliff1990connectionist, rebuffi2017icarl} methods. 
Regularization-based methods can be viewed as a way to shield the weights and therefore the learned representations for previous tasks from interference while learning new tasks. However, these methods are often overly focused on previous tasks, and the limited capacity of Deep Neural Networks (DNNs) makes them inflexible for learning new tasks~\cite{parisi2019continual}.
Rehearsal-based methods are more popular in the literature due to their simplicity and superior performance~\cite{buzzega2021rethinking, cha2021co2l}. However, they lack a mechanism to explicitly tackle representation drift~\cite{caccia2022new}. Parameter-isolation methods allocate a distinct set of parameters for new tasks, but they become memory-intensive as the number of tasks increases.
Overall, while each category of methods has its own advantages and disadvantages, none of them fully address representation drift efficiently.
Therefore, novel strategies are required to effectively tackle this challenge, upholding performance across new and old tasks.

\textbf{Representation Drift.}
In the context of CL, representation drift is a phenomenon in which previously learned representation clusters tend to drift while learning new tasks. \citet{murata2020happening} propose a representation-based evaluation framework to evaluate the impact of representation drift by freezing different layers after CL and retraining the remaining layers on all tasks. \citet{caccia2022new} propose two loss functions to mitigate representation drift in online class incremental learning (CIL) by learning new task samples separate from the buffered task samples. However, these methods are not directly applicable to DIL, where there is no separation between the seen and the new classes. To address this issue, \citet{yu2020semantic} use metric learning to mitigate feature drift in CIL, while \citet{zhang2022feature} propose a framework to quantify feature forgetting in CL and learn separate task-wise adapters to combat feature forgetting. However, these approaches entail memory overhead, which grows with the number of tasks. 


\begin{figure*}[t]
    \centering
    \includegraphics[width=0.99\textwidth]{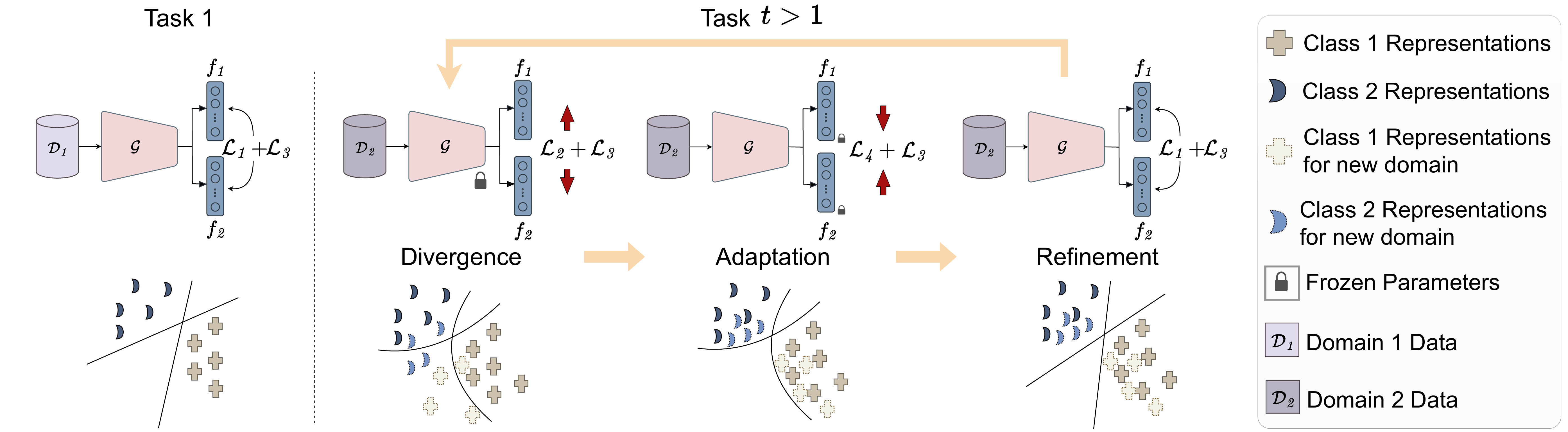}
    \caption{Our proposed method, \textit{DARE}, assimilates the knowledge about the new task while preserving the representations from earlier tasks by adopting a three-stage learning process in DIL. In the first two stages, \textit{Divergence} and \textit{Adaptation}, the model learns the representations of new domains within the cluster of old ones (rather than the other way around, which can exacerbate catastrophic forgetting). The final stage, \textit{Refinement}, helps the model learn the new domain samples.}
    \label{fig:method}
\end{figure*}

\textbf{Domain Adaptation.}
Domain Adaptation (DA) aims to transfer knowledge learned from a source dataset to a target dataset with a related domain. \citet{saito2018maximum} propose a dual classifier setup for DA. The training process alternates between maximizing the discrepancy between classifiers for out-of-domain samples and minimizing the discrepancy for in-domain samples by freezing the classifiers and learning the feature encoder to predict the same logits in both classifiers. In this way, the model adapts the representation of new domains in the space of old ones. Subsequently, \citet{yang2021multiple} proposed an extension with more than two classifiers to improve accuracy. Following the dual classifier approach, \citet{lv2022causality} propose a causally inspired framework with dual classifiers for DA. This framework aims to learn domain-independent representations in the encoder and to learn dual classifiers on complementary features using an adversarial mask. While DA focuses on the forward transfer of knowledge from old tasks to new tasks, DIL focuses on both forward and backward transfers, where the model must retain the knowledge of old tasks.


\section{Methodology}

Domain incremental learning (DIL) involves a sequence of $T$ tasks that become progressively available over time, with each task representing a shift in the input data distribution while the classes remain constant across tasks. During each task $t \in \{1, 2, .., T\}$, samples and their corresponding labels $\{(x_i, y_i)\}_{i=1}^{N}$ are drawn from the task-specific distribution $\mathcal{D}_t$~\cite{van2019three}. The CL model is optimized sequentially on each domain, and inference is carried out on all domains seen so far. An optimal CL model would learn to predict new distributions of input samples while retaining its knowledge of the initial tasks.


\subsection{Proposed Method - DARE}

Our method aims to enable efficient DIL by mitigating the abrupt representation drift at the task boundaries and adapting the learned representations to consolidate new information in a manner that reduces interference. To this end, we propose \textit{DARE} which employs a three-stage (\textit{Divergence}, \textit{Adaptation}, and \textit{REfinement}) learning mechanism that encourages the model to learn the representations of the samples belonging to the new task in the subspace spanned by the representations of the old tasks. This allows the model to consolidate new information without considerably disrupting the representations of the old tasks and adapt the decision boundary for the consolidated representations.

DARE utilizes an encoder $g$ to extract semantically meaningful representations from the input image, and dual classifiers $f_1$ and $f_2$ to project these representations to the class distribution (as depicted in Figure \ref{fig:method}). To learn more general and robust representations, and enforce the two classifiers to have different decision mechanisms, we employ the cross-entropy loss in the first classifier $f_1$ and the supervised contrastive loss~\cite{khosla2020supervised} in the second classifier, $f_2$. This equips our method with multiple viewpoints of the input data and ensures that the two classifiers sufficiently diverge. Furthermore, supervised contrastive loss provides the benefit of learning generalizable features across different domains while facilitating the learning of discriminative features across different classes~\cite{cha2021co2l}. Hence, the two learning objectives complement each other. 

In addition, we employ a buffer with bounded memory in which we store a portion of the learning task samples, labels, and logits from both classifiers. To this end, we propose an effective buffer sampling strategy, called ``\textit{Intermediary Reservoir Sampling}" strategy (see Section \ref{dark_logits}) throughout the training process to sample from the current task data and store them in a buffer. This way, the memory buffer contains samples from past tasks that are replayed later during the training process. 


Concretely, the first task is learned with the combination of cross-entropy loss on $f_1$ and supervised contrastive loss on $f_2$ on shared representations, where $z_i=g(x_i)$:
\begin{align}
\begin{split}
\label{eq:ce}
 \mathcal{L}_{1} \triangleq  \displaystyle \mathop{\mathbb{E}}_{(x_{i}, y_{i}) \sim \mathcal{D}_{t}} [ \: \mathcal{L}_{ce} (f_1(z_i), y_i) + \mathcal{L}_{sup} (f_2(z_i), y_i) ] 
\end{split}
\end{align}
For subsequent tasks, learning unfolds in three stages that enable CL by helping the model effectively adapt to new tasks while preserving prior knowledge.

\subsubsection{Divergence}
The divergence stage aims to tighten the decision boundaries of the two classifiers around the representation space spanned by the samples of already learned tasks. This involves maximizing the divergence between the two classifiers so that they can identify incoming samples from the new tasks whose representations do not lie in the space spanned by the learned representations. 

Specifically, we fix the parameters of the encoder, $g$, and maximize the distance between the $\ell_2$-normalized logits predicted by $f_1$ and $f_2$. The discrepancy loss~\cite{tan2022hyperspherical} measures the disparity in the distributions of pairwise distances of the classifier outputs $f_1(z)$ and $f_2(z)$. Let $d_1$ be the pairwise distance between the $\ell_2$-normalized logits predicted by classifier $f_1$ for a batch of input samples $\mathcal{X}$:
\begin{align}
d_1 = \lVert f_1(z_i) - f_1(z_j) \rVert
\end{align}
where $\lVert . \rVert$ denotes Euclidean distance.
The similarity metrics $p(.)$ can then be modeled as a normal distribution:
\begin{align}
p(d_1) = C_1 \frac{1}{\sigma_{1}\sqrt{2\pi}}exp \left[ -\frac{1}{2} \frac{(d_1 - \mu_1)^2}{\sigma_1^2} \right] 
\end{align}
where $C_1$ is a constant and $\mu_1$, and $\sigma_1^2$ are set to $0$ and $\frac{1}{2}$ following~\cite{tan2022hyperspherical}. 
Letting $d_2$ be the pairwise distance between the $\ell_2$-normalized logits predicted by the classifier $f_2$ and $q(.)$ be the corresponding similarity metrics, the discrepancy loss defined as:
\begin{equation}
\begin{split}
\label{eq:discrepancy}
\mathcal{L}_2 \triangleq \mathop{\mathbb{E}}_{\mathcal{X} \sim D_{t}} [p(d_1) \text{log} q(d_2) + (1-p(d_1))\text{log}(1-q(d_2))]
\end{split}
\end{equation}
%
Intuitively, as illustrated in Figure \ref{fig:method}, the divergence stage tightens the decision boundaries around the previous task samples by forcing the classifiers to maximize the distance in the predictions for the samples from the new task, while maintaining correct predictions on the previous tasks. This is achieved by minimizing the cross-entropy loss ($f_1$), supervised contrastive loss ($f_2$), and a consistency loss on the buffer samples in addition to maximizing the discrepancy loss on the new task samples:
\begin{align}
\begin{split}
\label{eq:consist}
 \mathcal{L}_{3} \triangleq 
 &\displaystyle \mathop{\mathbb{E}}_{(x_{i}', y_{i}', \zeta_{1,2}') \sim D_{m}} \big[ \alpha  \lVert \zeta_{1,2}' - f_{1,2}(z_i')\rVert_{2} \\
 &+ \mathcal{L}_{ce} (f_1(z_i'), y_i) + \mathcal{L}_{sup} (f_2(z_i'), y_i) \big]
\end{split}
\end{align}
%
where $\zeta_1$ and $\zeta_2$ are $f_1$ and $f_2$'s saved logits in the buffer, and $\alpha$ is a weighting parameter. The consistency loss encourages the classifiers to adapt their decision boundaries while maintaining the semantic relationships between the classes and enforces them to remain close to the optimal solution found for previous task samples in the memory buffer. Hence, the overall loss for the divergence stage is given by $\mathcal{L}_{\mathcal{D}} = \mathcal{L}_{2} + \mathcal{L}_{3}$.



\subsubsection{Adaptation}
The Divergence stage is followed by the Adaptation stage, which aims to adapt the encoder $g$ so that the representations of the new task samples are adapted within the subspace spanned by the already learned representations of previous tasks. Hence, the goal is to learn a consolidated representation space that supports the samples of the new tasks while remaining close to the optimal representations for the previously learned tasks. This is achieved by freezing the classifiers $f_1$ and $f_2$, and minimizing the discrepancy between their predictions. This enforces the encoder $g$ to adapt the representations so that the two classifiers agree on their predictions. The corresponding minimization loss for the discrepancy between $f_1$ and $f_2$ is given by;
\begin{equation}
\begin{split}
\label{eq:min}
\mathcal{L}_4 \triangleq -\mathop{\mathbb{E}}_{\mathcal{X} \sim D_{t}} [ p(d_1) \text{log} q(d_2) + (1-p(d_1))\text{log}(1-q(d_2))]
\end{split}
\end{equation}
%
%
%
Divergence and Adaptation can also be interpreted as a form of adversarial learning in which, first, the discriminators $f_1$ and $f_2$ are trained to discriminate the samples of the new task from those belonging to the old tasks by maximizing the discrepancy between the classifiers. Consequently, the generator, $g$, is trained to deceive the discriminators by extracting features that minimize the discrepancy between the two classifiers. Thus, during the course of learning the new task, the representations of the new task samples are gradually adapted to lie within the support spanned by the representations of previously learned tasks, rather than the other way around, which effectively reduces the drift in representation, and hence mitigates forgetting. The total loss for this stage is given by $\mathcal{L}_{\mathcal{A}} = \mathcal{L}_{3} + \mathcal{L}_{4}$. 

\subsubsection{Refinement}
Finally, the Refinement stage aims to refine the encoder and classifiers to effectively consolidate the new task information with the previously learned knowledge such that a learned consolidated representation space and decision boundary perform well for all the tasks seen so far. This involves training the encoder $g$, and the classifiers $f_1$ and $f_2$ to predict the correct classes for new task samples, while also minimizing a consistency loss with respect to the stored samples in the buffer. This encourages the model to learn the new task while maintaining previously acquired knowledge. The loss used at this stage is $\mathcal{L}_{\mathcal{R}} = \mathcal{L}_{1} + \mathcal{L}_{3}$.

Note that we iterate through the three stages multiple times while learning each task. This enables the model to gradually adapt the representations and decision boundary to acquire and consolidate information from the new task while mitigating the drift in representations and hence forgetting. Our proposed method is detailed in Algorithm \ref{alg:method}. 

\subsection{Intermediary Reservoir Sampling}
\label{dark_logits}
The proposed method to populate the replay buffer in DER utilizes Reservoir Sampling \cite{vitter1985random}. However, this uniform distribution throughout the learning trajectory does not optimize the storage of the exemplars. There is a nontrivial probability that logits are stored in the buffer at the very beginning or end of learning a task, leading to suboptimal performance. To improve this, we propose the ``\textit{Intermediary Reservoir Sampling (IRS)}" strategy, which employs a normal distribution over the learning trajectory of each task. The mean of the distribution is set to the intermediate stages, and the buffer is populated accordingly. This incentivizes the storage of logits with more "dark knowledge" about the current task, which in turn propagates the knowledge across future tasks through distillation. This approach aligns with recent research in knowledge distillation~\cite{wang2022efficient}, which suggests distilling with respect to an intermediate teacher model to capture maximum information on the ``dark knowledge" between data samples. 
We defer Algorithm \ref{alg:dark_logit} and ablation studies for IRS (Table \ref{tab:sampling_ablation}) to Appendix.

\section{Experimental Setup}

We address the issue of sudden changes in data representation that occur with the introduction of new domains. To tackle this problem, we propose a novel approach to mitigate drift, and our results demonstrate that addressing this issue leads to improved performance on standard DIL benchmarks. 
To perform our experiments, we use the \emph{mammoth} framework~\cite{buzzega2020dark} to emulate DIL scenarios and implement our approach on top of the ResNet-18 architecture~\cite{he2016deep}, following previous works~\cite{buzzega2020dark, rebuffi2017icarl}. We modify the network to include our proposed approach, in which the encoder $g$ retains the default ResNet-18 structure, and the classification heads $f_1$ and $f_2$ are linear layers projecting the encoded representations from $g$ to a number of classes $C$, such that $f_{1,2}: \mathbb{R}^{d} \to \mathbb{R}^{C}$, where $d$ is the dimension of flattened representations from the encoder. We train our method with a batch size of 32, for 50 epochs per task on all datasets. 

\begin{algorithm}[!h]
\centering
\caption{Learning Algorithm for DARE}
\label{alg:method}
\begin{algorithmic}
    \STATE \textbf{input:} Data streams $\mathcal{D}_{t}$, model with backbone $g$ and two classifiers $f_1$, $f_2$ parameterized by $\theta$, memory buffer $\mathcal{M} = \{\}$
    
    \FORALL {tasks $t \in \{1, 2,..,T\}$}
        \FOR{epochs $e \in \{1, 2,..,E\}$} 
            \IF{$t=1$}
                \FOR{batch ${(x_t, y_t)} \in \mathcal{D}_{t}$}
                    \STATE Compute $\mathcal{L}_{1}$ 
                    \STATE Update the model based on $\nabla_\theta \mathcal{L}_{1}$
                \ENDFOR
            \ELSE
                \IF{$e\%3==0$}
                \STATE \hfill $\triangleright$ Divergence
                    \FOR{batch ${(x_t, y_t)} \in \mathcal{D}_{t}$} 
                        \STATE Freeze the encoder $g(.)$
                        \STATE $\zeta_1, \zeta_2 = f_1(x_t), f_2(x_t)$                    
                        \STATE Sample batch ${(x', y', \zeta_{1,2}')} \in \mathcal{M}$ 
                        \STATE Compute $\mathcal{L}_{\mathcal{D}} = \mathcal{L}_{3} + \mathcal{L}_{2}$
                        \STATE Update classifiers based on $\nabla_\theta \mathcal{L}_{\mathcal{D}}$
                    \ENDFOR
                \ELSIF{$e\%3==1$}
                    \STATE \hfill $\triangleright$ Adaptation
                    \STATE Unfreeze the model
                    \FOR{batch ${(x_t, y_t)} \in \mathcal{D}_{t}$}
                        \STATE Freeze the classifiers $f_1, f_2$
                        \STATE $\zeta_1, \zeta_2 = f_1(x_t), f_2(x_t)$ 
                        \STATE Sample batch ${(x', y', \zeta_{1,2}')} \in \mathcal{M}$ 
                        \STATE Compute $\mathcal{L}_{\mathcal{A}} = \mathcal{L}_{3} + \mathcal{L}_{4}$
                        \STATE Update the backbone based on $\nabla_\theta \mathcal{L}_{\mathcal{A}}$
                    \ENDFOR
                \ELSIF{$e\%3==2$}
                    \STATE \hfill $\triangleright$ Refinement
                    \STATE Unfreeze the model
                    \FOR{batch ${(x_t, y_t)} \in \mathcal{D}_{t}$}
                        \STATE $\zeta_1, \zeta_2 = f_1(x_t), f_2(x_t)$ 
                        \STATE Sample batch ${(x', y', \zeta'_{1,2})} \in \mathcal{M}$
                        \STATE Compute $\mathcal{L}_{\mathcal{R}} = \mathcal{L}_{1} + \mathcal{L}_{3}$
                        \STATE Update the model based on $\nabla_\theta \mathcal{L}_{\mathcal{R}}$
                    \ENDFOR 
                \ENDIF
            \ENDIF
        \STATE Update $\mathcal{M} \leftarrow \textit{IRS} (x, y, \zeta_{1,2})$ \hfill $\triangleright$ Algorithm \ref{alg:dark_logit}
        \ENDFOR
    \ENDFOR
    \STATE \textbf{return: }{model $\theta$}
\end{algorithmic}
\end{algorithm}

We evaluate our proposed method in DIL setting~\cite{van2019three} on two diverse datasets. \textbf{DN4IL} (DomainNet for Domain-IL) is a challenging dataset consisting of six vastly diverse domains and samples belonging to 100 classes \cite{gowda2023a}. On the other hand, \textbf{iCIFAR-20} \cite{xie2022general} is the DIL setup of the CIFAR-100 dataset~\cite{krizhevsky2009learning}, where the 20 supercategories are considered actual classes and the five subcategories are considered new domains. We focus on evaluating models on datasets that closely mimic real-world domain shifts, as opposed to the transformed versions of MNIST commonly used in the literature. 
More information about the datasets and training is deferred to Appendix.

\begin{table*}[t]
\centering
\caption{Results on DIL benchmarks learned with varying buffer sizes, averaged over 3 runs. 
Accuracy determines the performance on all tasks learned by the model, and backward transfer (BWT) quantifies the degree to which learning a new task improves performance on previously learned tasks. \#P denotes the total count of trainable parameters (expressed in millions). \protect\footnotemark}
\label{tab:main}
\begin{tabular}{cl|ccc|ccc}
\hline
\multirow{2}{*}{\begin{tabular}[c]{@{}c@{}}Buffer\\ Size\end{tabular}} & \multirow{2}{*}{Method} & \multicolumn{3}{c|}{iCIFAR-20} & \multicolumn{3}{c}{DN4IL} \\ \cline{3-8}
& & \#P $\downarrow$ & BWT $\uparrow$ & Last Accuracy $\uparrow$ & \#P $\downarrow$ & BWT $\uparrow$ & Last Accuracy $\uparrow$ \\
\hline
\multirow{2}{*}{-} & Joint & 11.18 & - & 79.61\tiny{$\pm$0.13} & 11.22 & -  & 59.93\tiny{$\pm$1.07} \\
 & SGD & 11.18 & -43.72\tiny{$\pm$1.07} & 49.40\tiny{$\pm$0.53} & 11.22 & -42.42\tiny{$\pm$0.00}  & 21.63\tiny{$\pm$0.42} \\
 \hline
 \multirow{6}{*}{50} & ER & 11.18 & -42.03\tiny{$\pm$0.27} & 50.23\tiny{$\pm$0.94} & 11.22  & -36.11\tiny{$\pm$0.26} & 24.24\tiny{$\pm$0.34} \\
 & DER++ & 11.18 & -40.63\tiny{$\pm$0.49} & 52.68\tiny{$\pm$1.10} & 11.22 & -29.05\tiny{$\pm$1.35} & 28.08\tiny{$\pm$0.99} \\
 & DARE & 11.19 & \textbf{-34.98}\tiny{$\pm$1.52} & \textbf{53.66}\tiny{$\pm$0.59} & 11.27 &  \textbf{-22.98}\tiny{$\pm$0.62} & \textbf{32.32}\tiny{$\pm$0.53} \\
 \cline{2-8}
 & CLS-ER & 33.57 & - & \textbf{63.01}\tiny{$\pm$0.80} &  33.81 & - & 37.90\tiny{$\pm$1.15} \\
 & DUCA & 33.57 & - & 61.48\tiny{$\pm$0.25} & 33.81 & - & 38.91\tiny{$\pm$2.12} \\
 & DARE++ & 22.38 & - & 62.43\tiny{$\pm$0.37} &  22.54  & - & \textbf{40.51}\tiny{$\pm$0.17} \\
 \hline
 \multirow{6}{*}{100} & ER & 11.18 & -41.88\tiny{$\pm$0.59} & 50.85\tiny{$\pm$0.73} &  11.22 & -35.28\tiny{$\pm$1.20} & 24.67\tiny{$\pm$0.86} \\
 & DER++ & 11.18 & -37.33\tiny{$\pm$1.47} & 55.32\tiny{$\pm$0.69} & 11.22 & -27.78\tiny{$\pm$0.90} & 32.06\tiny{$\pm$1.05} \\
 & DARE & 11.19 & \textbf{-33.20}\tiny{$\pm$0.09} & \textbf{56.01}\tiny{$\pm$0.22} & 11.27 & \textbf{-19.37}\tiny{$\pm$0.43} & \textbf{37.16}\tiny{$\pm$0.62} \\
  \cline{2-8}
 & CLS-ER & 33.57 & - & 64.31\tiny{$\pm$0.43} &  33.81 & - & 39.30\tiny{$\pm$0.74} \\
 & DUCA & 33.57 & - & 62.59\tiny{$\pm$0.27} & 33.81 & - & 43.09\tiny{$\pm$0.14} \\
 & DARE++ & 22.38 & - & \textbf{64.59}\tiny{$\pm$0.24} & 22.54  & - & \textbf{43.27}\tiny{$\pm$0.37} \\
 \hline
\multirow{6}{*}{200} & ER & 11.18 & -38.98\tiny{$\pm$0.74} & 52.57\tiny{$\pm$0.79} & 11.22  & -32.35\tiny{$\pm$0.51} & 27.45\tiny{$\pm$0.94} \\
 & DER++ & 11.18 & -33.61\tiny{$\pm$0.64} & 58.39\tiny{$\pm$0.38} & 11.22 & -23.99\tiny{$\pm$0.74} & 35.74\tiny{$\pm$0.67} \\
 & DARE & 11.19 & \textbf{-30.22}\tiny{$\pm$1.84} & \textbf{58.53}\tiny{$\pm$1.25} & 11.27 & \textbf{-14.69}\tiny{$\pm$0.19}  & \textbf{40.59}\tiny{$\pm$0.73} \\
  \cline{2-8} 
  & CLS-ER & 33.57 & - & \textbf{66.40}\tiny{$\pm$0.81
 } & 33.81  & - & 41.70\tiny{$\pm$1.41} \\
 & DUCA & 33.57 & - & 66.04\tiny{$\pm$0.36} & 33.81 & -  & \textbf{44.45}\tiny{$\pm$0.18} \\
 & DARE++ & 22.38 & - & 65.79\tiny{$\pm$0.92} & 22.54  & - & 44.11\tiny{$\pm$0.98} \\
\hline
\end{tabular}
\end{table*}


\section{Empirical Results}

We compare our approach with state-of-the-art rehearsal-based methods in CL literature under uniform experimental settings, focusing on the challenging low buffer regime where representation drift is most pronounced~\cite{caccia2022new}. For a comprehensive study, we selected standard methods such as ER~\cite{riemer2018learning}, DER++ \cite{buzzega2020dark}, CLS-ER \cite{arani2022learning}, and DUCA~\cite{gowda2023a}. CLS-ER uses slow and fast learners to distill knowledge from past tasks, while DUCA is a multimemory system that integrates shape cognitive bias. To consolidate learned knowledge, we employed a semantic memory, an exponential moving average (EMA) of the learning model, comparing it with CLS-ER and DUCA. Our proposed method is 'DARE,' and 'DARE++' represents the results of the EMA model in an extended dual-memory version. We also report both the upper bound, denoted \textit{Joint}, where training uses the entire dataset, and the lower bound, denoted \textit{SGD}, where training progresses through new domains without an additional buffer.

Table \ref{tab:main} presents the performance of DARE and other baselines on DIL benchmarks. The results indicate consistent improvements in final accuracy (over all seen tasks) and backward transfer (BWT) when using a single learning model (DARE) with different buffer sizes. Additionally, DARE++ achieves comparable or even better results than other multi-memory based approaches. Notably, CLS-ER and DUCA require storing all multi-memory models in the device, leading to high memory requirements reflected in the number of parameters in the framework. The efficacy of DARE++ is evident from its performance on par with other multi-memory systems with significantly lower parameters.

\footnotetext{BWT numbers for methods that include an EMA model are not mentioned due to the stochastic nature of the EMA update.}

In the challenging scenario where the buffer size is limited to 50 in DN4IL, our proposed method, DARE, demonstrates significant improvements of 33.3\% and 15.1\% in accuracy over ER and DER++, respectively. Additionally, DARE trained with a smaller buffer outperforms the DER++ counterparts trained on larger buffer sizes. Similarly, DARE++ trained with a smaller buffer size outperforms CLS-ER trained with a larger buffer size. DN4IL is a highly challenging DIL dataset with significant domain shifts, and these improvements demonstrate the effectiveness of DARE.

\begin{figure*}[tbh]
\begin{center}
\begin{tabular}{ccc}
\includegraphics[width=0.29\textwidth]{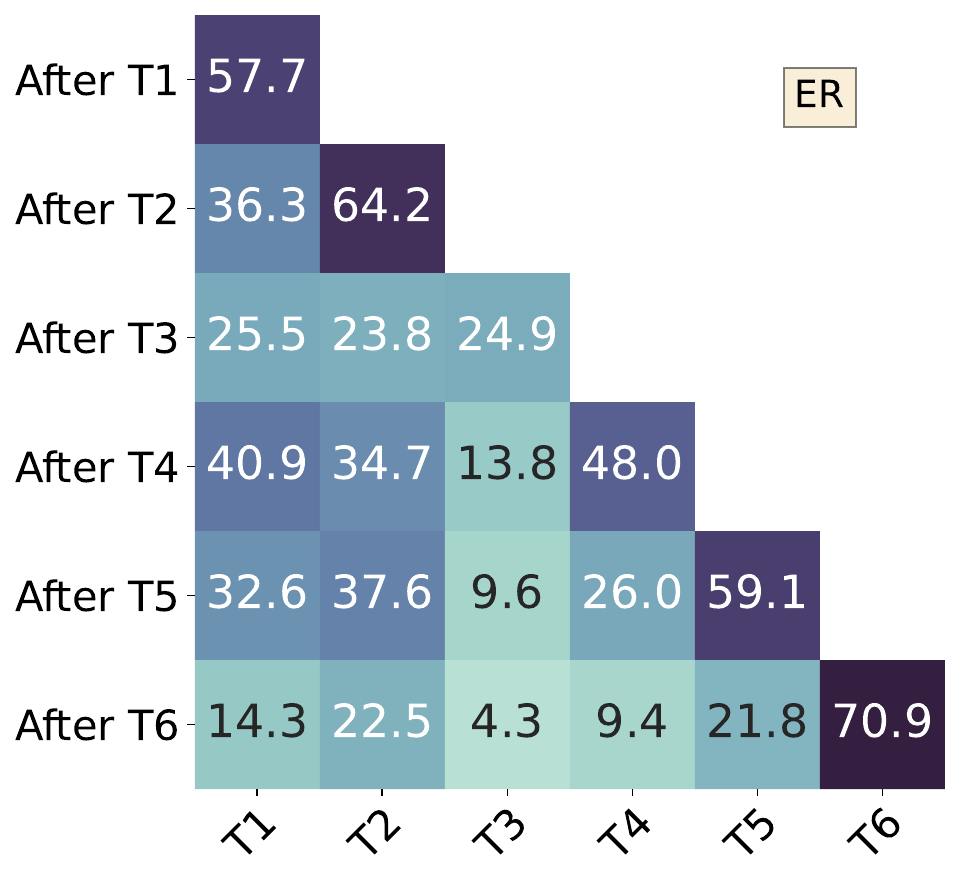} & 
\includegraphics[width=0.27\textwidth]{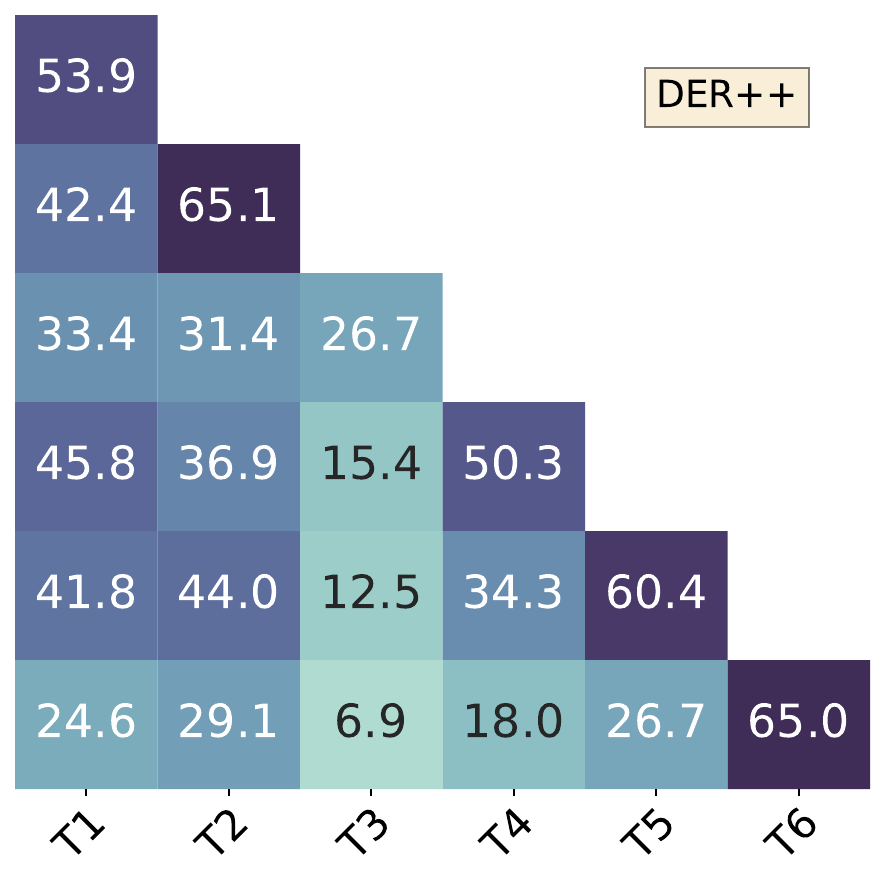} &
\includegraphics[width=0.27\textwidth]{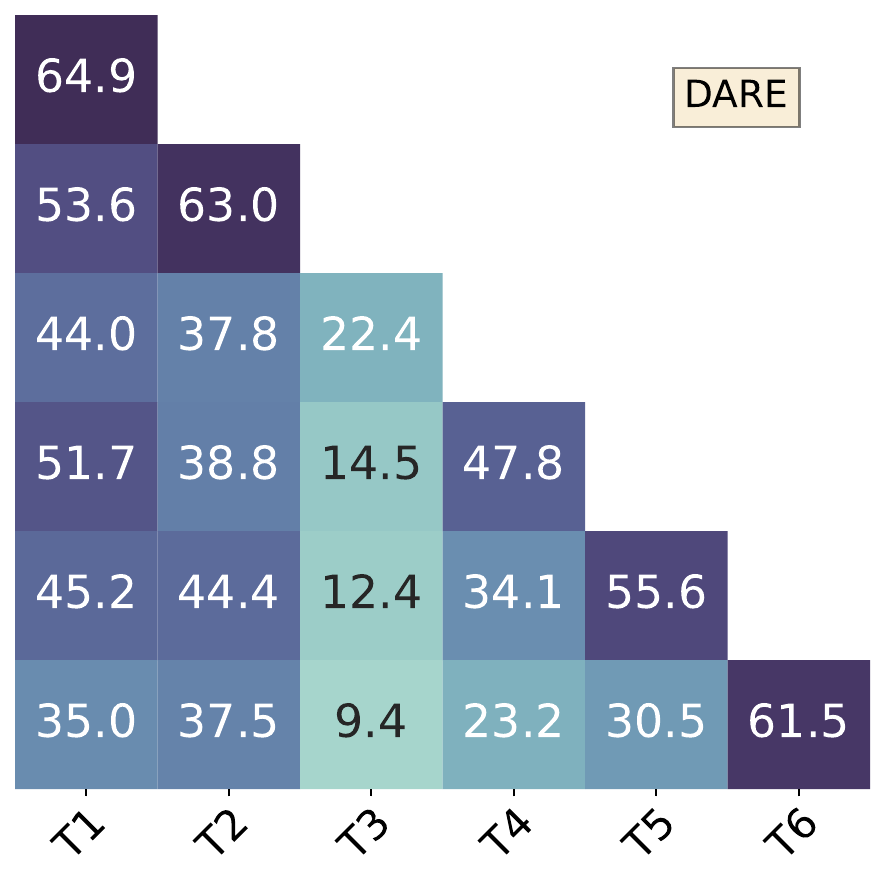}
\end{tabular}
\caption{Task-wise accuracy of different CL models while learning new tasks with buffer size 50. DARE retains more performance on seen domains compared to ER and DER++.}
\label{fig:task_heatmap}
\end{center}
\end{figure*}

\begin{figure*}[t]
\begin{center}
\begin{tabular}{cc}
\includegraphics[width=0.445\textwidth]{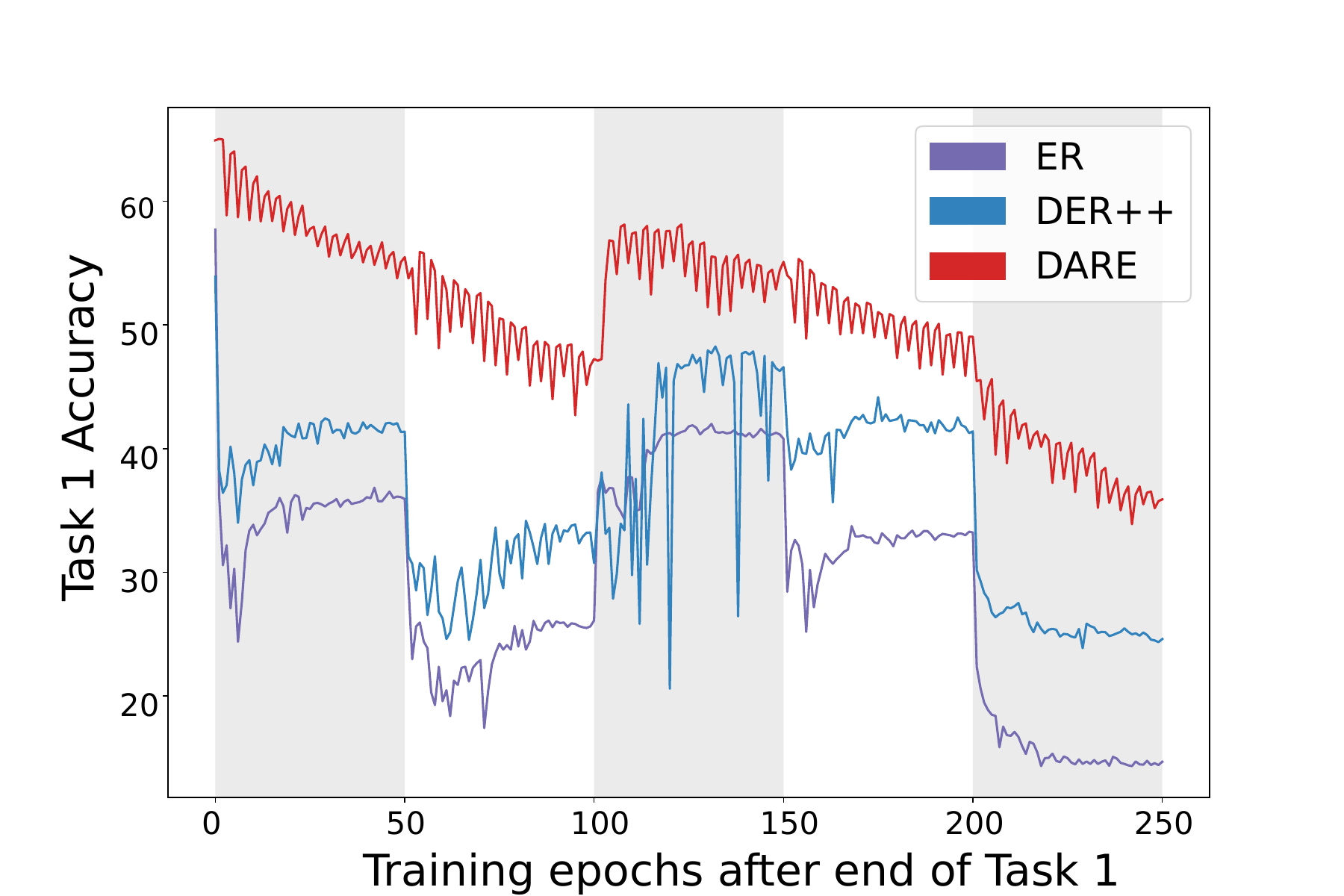} &
\includegraphics[width=0.445\textwidth]{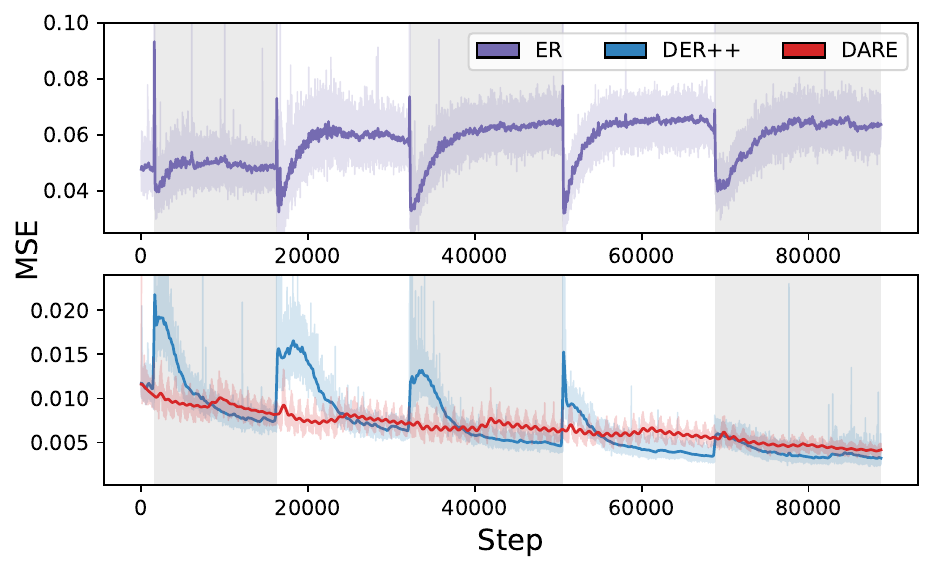} 
\end{tabular}
\caption{\textbf{Representation drift analysis}. Left: Epoch-wise accuracy on Task 1 samples, while learning future tasks (shaded regions indicate new tasks). Right: Iteration-wise drifts for buffered samples for CL methods trained with a buffer size of 50. It is evident that DARE effectively reduces representation drift compared to other methods.}
\label{fig:iteration_drift}
\end{center}
\end{figure*}

\begin{figure*}[t]
\begin{center}
\begin{tabular}{cc}
\includegraphics[width=0.34\textwidth]{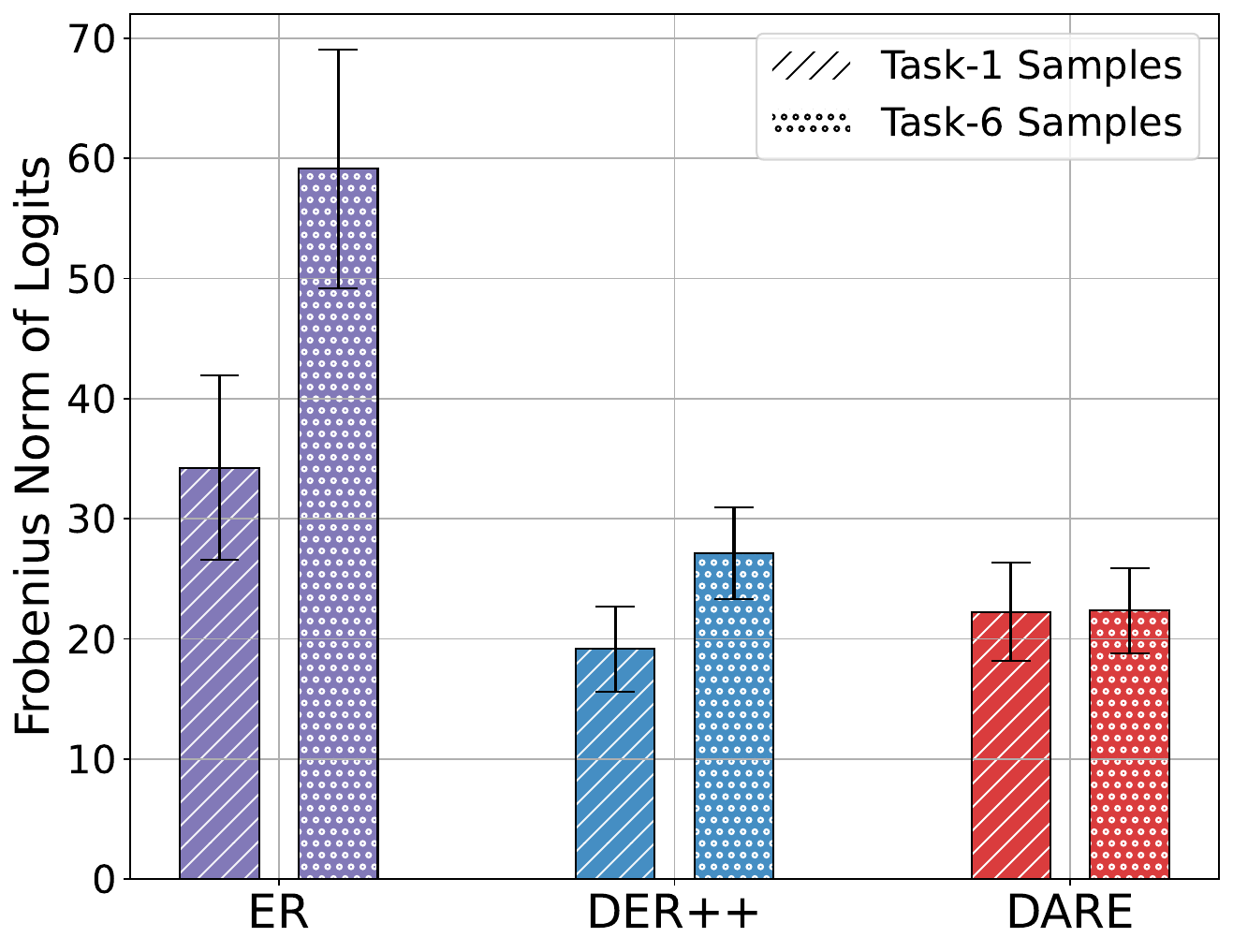} & 
\includegraphics[width=0.49\textwidth]{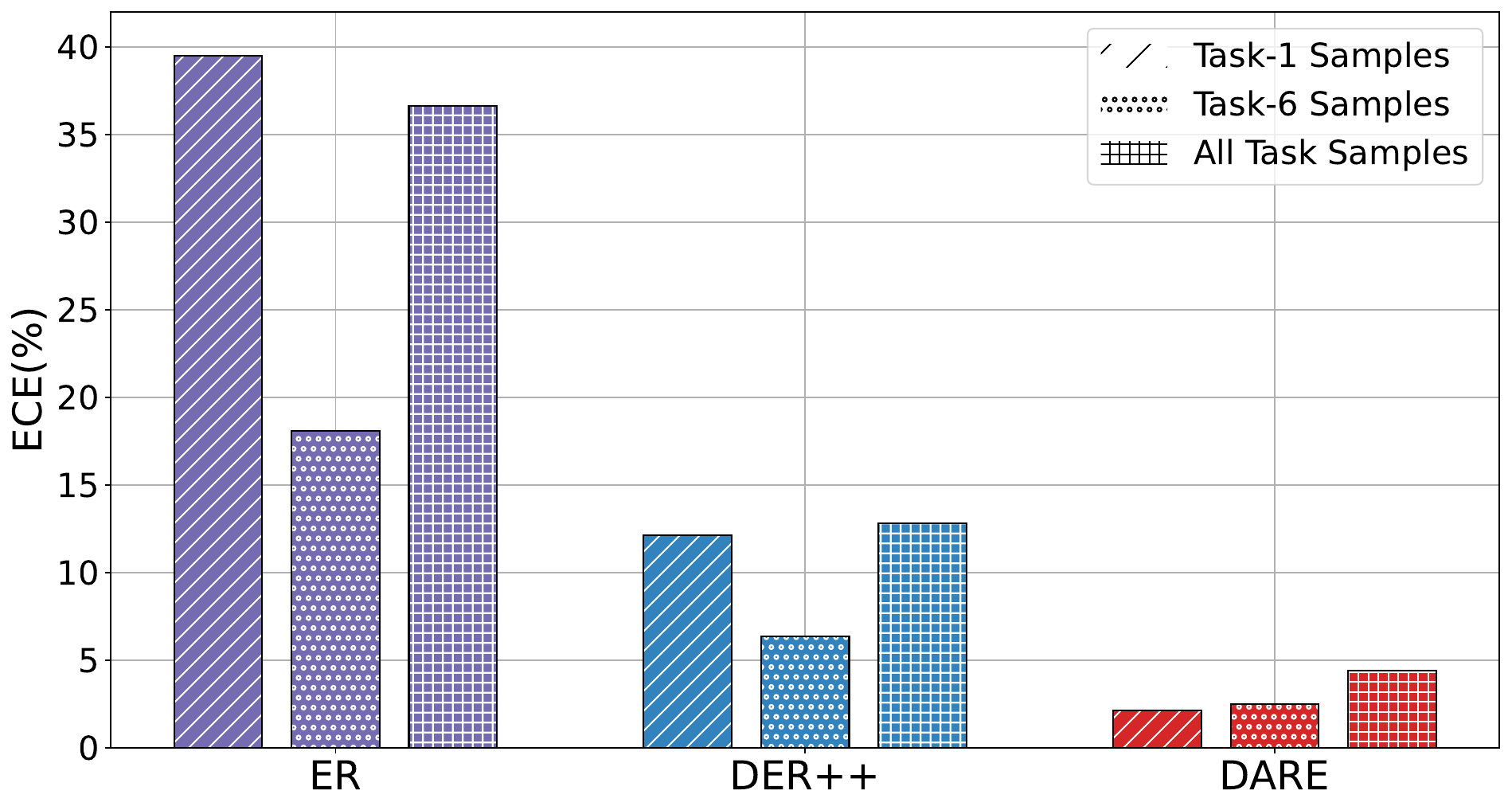} 
\end{tabular}
\caption{Model calibration and task recency bias analyses of different CL approaches learned with buffer size 200. Left: Logit norm analysis shows that DARE predicts logits with magnitudes smaller than DER++ (less overconfident) for recent task samples. Right: DARE has a lower calibration error compared to DER++ on samples belonging to different tasks.}
\label{fig:calibration}
\end{center}
\end{figure*}

Our results clearly indicate that DARE can effectively learn new domains while maintaining high performance on old tasks, even under complex and memory-restrictive settings. 
This can be attributed to our learning algorithm preserving representations of old tasks while acquiring new ones.

\section{Model Analysis}

We evaluate the effectiveness of our proposed approach in challenging scenarios through various analysis experiments, comparing its performance to single-model approaches.

\subsection{Task-wise Performance}

The extreme difference between the domains in every task warrants the study of the model's knowledge about the seen tasks while learning new tasks. Figure \ref{fig:task_heatmap} shows the task-wise accuracy of different CL approaches while learning new tasks. DARE retains the accuracy of old tasks better compared to ER and DER++. Furthermore, learning task 4 helps improve performance on task 1 across all CL algorithms, and this can be attributed to the similar nature of task 4 (painting) to task 1 (real). It is worth noting that DARE achieves higher performance in task 1 compared to DER++ which employs almost a similar learning algorithm except for the proposed IRS strategy. The consistency loss with respect to intermediate-stage checkpoints helps to learn the first task better than other approaches.


\subsection{Study of Representation Drift}

The representations learned for previous tasks in the backbone denote the knowledge of the model about the relationship between the input samples and the labels drawn from the data distribution of previous tasks. 
Modification to important weights in the network for the previous task while learning the new task is deemed to contribute to catastrophic forgetting in CL~\cite{mccloskey1989catastrophic}. 
Thus, analyzing the change in the representations of past tasks would shed some light on the amount of catastrophic forgetting. 

We analyze the representation drift of early task samples in two ways. First, we plot the accuracy of the task 1 validation set over the course of training in the other domains in Figure \ref{fig:iteration_drift} (left). This represents the disruptive nature of representation drift at task boundaries and their detrimental effect on the accuracy of seen tasks. It can be seen that both ER and DER++ undergo a significant decrease in performance for task 1 samples at the beginning of learning task 2. The same behavior is observed at the beginning of tasks 3, 5, and 6. However, DARE prevents such loss inaccuracy for task 1 samples while learning new tasks, as it inhibits disruptive updates to learned representation by design. 
We observe an increase in task 1 accuracy at the beginning of task 4 in all methods. 
This is explained by the backward transfer of task 4 (painting), which has features similar to task 1 (real) compared to other domains. DARE is flexible enough to allow for such a backward transfer of knowledge.

Second, we examine the iteration-wise representation drift of the buffered samples in Figure \ref{fig:iteration_drift} (right). The plot reveals that buffered sample representations experience a sudden shift at task boundaries (indicated by alternating shaded regions). However, DARE, with and without IRS, doesn't exhibit a comparable representation change. As we adjust future task representations into the initial task's representation space, the drift is minimal and gradual, contributing to reduced forgetting. Preventing harmful alterations in representations for samples from the same class set but different domains supports accurate sample classification.

\subsection{Task Recency and Model Calibration}

Task recency bias is an important concern in CL, wherein model predictions are biased more towards recent tasks and result in more forgetting for earlier task samples~\cite{wu2019large}.
In DIL, task recency bias can have severe consequences, particularly in safety-critical applications such as autonomous driving, where forgetting knowledge about old tasks can lead to misclassifications. 
Therefore, it is imperative to develop effective strategies to evaluate and address task recency bias in DIL, especially to ensure the reliability and safety of deployed models.

Although task recency bias has been extensively studied in CIL~\cite{wu2019large, masana2022class, hou2019learning, arani2022learning} and is straightforward to analyze as classes are distinct between different tasks, it has not been widely studied in the DIL scenario due to its inherent aspect where all tasks share the same set of classes.
As a step forward in studying this bias in DIL, we analyze the logit norms of different CL approaches. 
DNNs that predict logits with a larger magnitude or norm directly translate into overconfident predictions~\cite{chrysakis2023online}, which in turn can indicate the bias of a model toward samples belonging to a certain task. 
Figure \ref{fig:calibration} (left) illustrates the logit norms predicted by different CL models on the first and last domain samples. 
While ER and DER++ are more confident on the last task samples compared to the samples belonging to the first task, DARE achieves more uniformly distributed confidence over old and new task samples.
Additionally, Figure \ref{fig:calibration} (right) illustrates the calibration error~\cite{guo2017calibration} of the model on the initial, last, and all task samples. 
It is further evident that training with DARE achieves a lower calibration error compared to other CL methods.

\begin{table}[tb]
\centering
\caption{Ablation study on the effect of loss components on the last accuracy of DARE, averaged over 3 runs.}
\label{tab:loss_ablation}
\begin{small}
\resizebox{\columnwidth}{!}{
\begin{tabular}{ccclll}
\toprule
\begin{tabular}{c} Buffer \\ size \end{tabular}
 & DARE & - $\mathcal{L}_1$ & \multicolumn{1}{c}{- $\mathcal{L}_2$} & \multicolumn{1}{c}{- $\mathcal{L}_3$} & \multicolumn{1}{c}{- $\mathcal{L}_4$} \\
\midrule
50 & \textbf{32.32}\tiny{$\pm$0.53} & 11.22\tiny{$\pm$2.32} & 31.11\tiny{$\pm$1.09} & 24.69\tiny{$\pm$0.34} & 31.63\tiny{$\pm$0.98} \\
200 & \textbf{40.59}\tiny{$\pm$0.73} & 22.26\tiny{$\pm$0.46} & 38.69\tiny{$\pm$0.29} & 24.55\tiny{$\pm$0.83} & 38.92\tiny{$\pm$0.48} \\
\bottomrule
\end{tabular}}
\end{small}
\end{table}

\subsection{Effectiveness of Individual Components}

The effectiveness of our method is demonstrated through an ablation study (Table \ref{tab:loss_ablation}), where three interconnected loss functions are crucial. Removing any loss function results in performance decline, especially after removing $\mathcal{L}_1$ and $\mathcal{L}_3$, impacting accuracies for both current and previous tasks. In particular, $\mathcal{L}_2$ and $\mathcal{L}_4$ contribute to gains of 3.89\% and 2.18\% for buffer size 50, maximizing Divergence and minimizing Adaptation steps. The synergy among the three stages—Divergence, Adaptation, and Refinement—optimizes the representation space for new tasks, ensuring optimal learning. Their interdependence is crucial; isolated analysis risks divergence, and the absence of $\mathcal{L}_3$ during Divergence leads to catastrophic forgetting. This approach also minimizes representation drift (see Figure \ref{fig:iteration_drift}).

\section{Conclusion}

We proposed a novel method to address representation drift in domain-incremental learning. Our proposed method, DARE, mitigates representation drift at task boundaries and effectively assimilates new domain information into the feature space of old task samples. The inclusion of an effective buffer sampling strategy allows the preservation of the dark knowledge learned on old tasks when learning new ones. Our empirical evaluation demonstrated that DARE outperforms existing methods across different DIL benchmarks, with less forgetting and improved performance on seen domains. Furthermore, DARE exhibits efficient memory and computational usage, reduces bias towards recent task samples, and inhibits abrupt representation drift at task boundaries. These results demonstrate DARE's efficacy and the potential for practical applications in continual learning.

\section*{Limitations and Future Work}

One particular area of focus for enhancement that we endeavor to tackle pertains to the enhancement of our methodology to lessen the reliance on task-id, which is presently vital for the IRS buffer sampling strategy. Nevertheless, we can integrate mechanisms for independently identifying task transitions, such as monitoring variations in loss metrics.

\bibliography{example_paper}
\bibliographystyle{icml2024}



\clearpage
\newpage

\appendix
\onecolumn
\section{Appendix}

\subsection{Evaluation Metrics}

To evaluate the performance of different models under different settings, we selected two main metrics widely used in the CL literature. We formalize each metric below.

\begin{enumerate}[noitemsep,topsep=0pt]
 \item \textbf{Last Accuracy} defines the final performance of the CL model on the validation set of all the tasks seen so far. Concretely, given that tasks are sampled from a set $t \in {1,2 ..., T}$, where $T$ is the total number of tasks and $a_{k,j}$ is the accuracy of a CL model on the validation set of the task $k$ after learning task {$j$}, last accuracy $A_{last}$ is as follows:
 \begin{equation}
A_{last} = \frac{1}{T} \sum_{k=1}^{T} a_{k, T}
 \end{equation}
 \item \textbf{Backward Transfer (BWT)} defines the influence of the learning task $t$ on previously seen tasks $k < t$. Positive BWT implies that the learning task $t$ increased performance on previous tasks, while negative BWT indicates that the learning task $t$ affected the performance of the model on previous tasks. Formally, BWT is as follows:
 \begin{equation}
BWT = \frac{1}{T-1} \sum_{j=1}^{T-1} a_{T, j} - a_{j, j}
 \end{equation}
\end{enumerate}

\subsection{Datasets}

\begin{figure}[tbh]
\begin{center}
\includegraphics[width=0.475\textwidth]{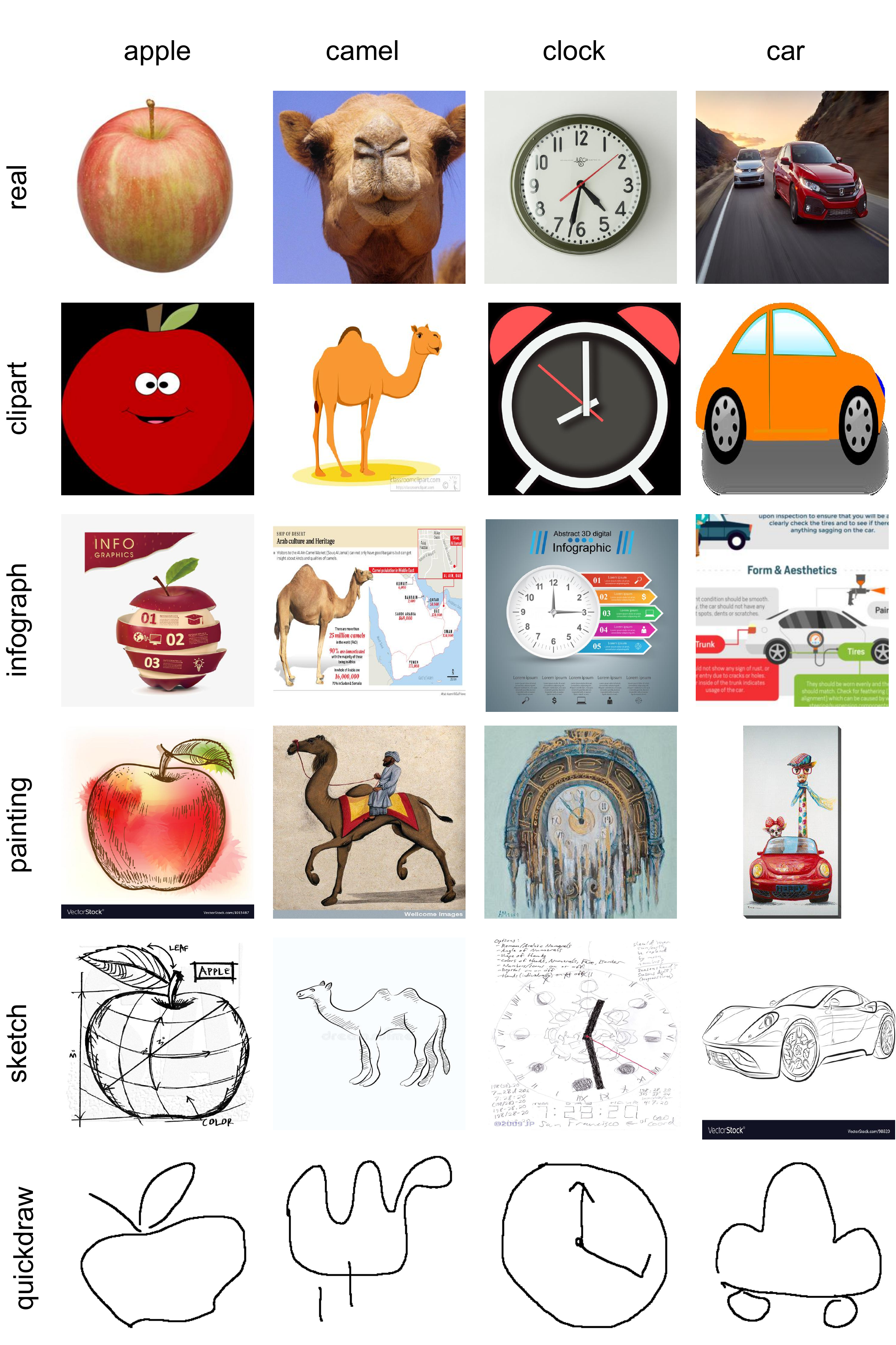}
\caption{Visualization of domain shifts in the DN4IL dataset.}
\label{fig:dn4il}
\end{center}
\end{figure}

\textbf{DN4IL}~\cite{gowda2023a} is a subset of the standard DomainNet dataset~\cite{peng2019moment} proposed for large-scale unsupervised domain adaptation composed of 345 categories. DN4IL is a class-balanced dataset with 20 supercategories and five classes under each supercategory. In total, the dataset consists of 100 categories spanning over 6 different domains namely, `sketch', `real', `quickdraw', `painting', `infograph', and `clipart' with approximately 67k training images and 19k test images of shape $64\times64$. Domain-incremental learning (DIL) scenario on DN4IL is more challenging compared to other datasets conventionally used for DIL as the distribution shift is more prominent. Figure \ref{fig:dn4il} shows some examples of different domains in the dataset.

\textbf{iCIFAR-20}~\cite{xie2022general} is the DIL version of CIFAR-100~\cite{krizhevsky2009learning} dataset. iCIFAR-20 is a class-balanced dataset with 20 supercategories and five classes under each supercategory. The 20 supercategories are considered as real labels, and the five subcategories under each label is considered as a new domain. The dataset consists of approximately 50k training images and 10k test images of size $32\times32$.

\subsection{IRS - Intermediary Reservoir Sampling}
Motivated by its effectiveness in Reinforcement Learning~\cite{rolnick2019experience}, rehearsal-based methods in continual learning settings store a subset of input samples/exemplars and their corresponding labels in the replay buffer and interleave them while learning new tasks. 
Ideally, the replay buffer is expected to model the data distribution of all previous tasks, and the training algorithm samples exemplars from the buffer and interleaves them with the current task samples while learning a new task, thus mitigating forgetting the knowledge of old tasks. 
Rehearsal-based methods are widely used in CL and different approaches have been proposed to populate the buffer~\cite{rebuffi2017icarl, chaudhry2018riemannian}. 

Dark Experience Replay (DER)~\cite{buzzega2020dark} proposes to store logits along with exemplars and to learn the model on new tasks while emulating their earlier responses to old task samples.
Analogous to logit replay, many works have tried distilling logits from a teacher model, typically a snapshot of the model at task boundaries~\cite{douillard2022dytox, li2017learning, DBLP:journals/corr/abs-1911-03462} or exponential moving average of the model~\cite{arani2022learning} to mitigate forgetting. 
Concretely, the regularization loss on the logits distills the `dark knowledge'~\cite{hinton2014dark} learned by the model in the previous tasks into the weights of the model being trained. This dark knowledge constitutes more information about the relationships among different classes of input, thus guiding the learning model better discriminate among samples belonging to different tasks and different classes as well. 
Thus, the information contained in the logits contributes significantly to the accuracy of the learning model on all the seen tasks.

\begin{table}[tbh]
\centering
\caption{Comparison of the reservoir sampling and the proposed IRS buffer sampling strategy. Results are on DN4IL dataset trained with buffer size 50 for six tasks.}
\label{tab:sampling_ablation}
\begin{small}
\begin{tabular}{l|l|cc}
\hline
Metric & Method & \begin{tabular}[c]{@{}c@{}}Reservoir\\ Sampling\end{tabular} & IRS \\
 \hline
\multirow{2}{*}{BWT} & DER++ & -23.99\tiny{$\pm$0.74} & -22.69\tiny{$\pm$3.71} \\
 & DARE & -15.77\tiny{$\pm$0.69} & \textbf{-14.69}\tiny{$\pm$0.19} \\
\hline
\multirow{2}{*}{Last Accuracy} & DER++ & 35.74\tiny{$\pm$0.67} & 37.60\tiny{$\pm$1.21} \\
 & DARE &  36.17\tiny{$\pm$0.38} & \textbf{40.59}\tiny{$\pm$0.73} \\
 \hline
\end{tabular}
\end{small}
\end{table}

\begin{algorithm}[t]
\centering
\caption{Intermediary Reservoir Sampling (IRS)}
\label{alg:dark_logit}
\begin{small}
\begin{algorithmic}
    \STATE \textbf{input: } Data streams $\mathcal{D}_{t} \forall \{ t = 1,...,T \}$,  model $f_\theta$, memory buffer $\mathcal{M}$ , number of seen examples $N$, input sample $x$, label $y$, logit $z$.
    
    \FORALL {tasks $t \in \{1, 2,..,T\}$}
      \FOR {epochs $ep \in \{1, 2,..,E\}$} 
        \FOR{minibatch $ \mathcal{B} \rightarrow {(x_t, y_t)} \in \mathcal{D}_{t}$ of size $|\mathcal{B}|$}
            \STATE $z_t = f_\theta(x_t)$
            \IF{uniform $[0,1] < \frac{1}{\sigma \sqrt{2\pi}} e^{{{ - \left( {ep - \frac{E}{2}} \right)^2 } \mathord{\left/ {\vphantom {{ - \left( {e - \mu } \right)^2 } {2\sigma ^2 }}} \right. \kern-\nulldelimiterspace} {2\sigma ^2 }}}$}
                \IF{$ |\mathcal{M}| < N$}
                    \STATE $\mathcal{M}[N] \leftarrow (x_t, y_t, z_t) $
                \ELSE
                    \STATE $i \sim [0,N]$  \hfill$\triangleright$ Random index
                    \IF{$i < |\mathcal{M}| $}
                        \STATE $\mathcal{M}[i] \leftarrow (x_t, y_t, z_t) $
                    \ENDIF
                \ENDIF
            \ENDIF
          $N$ += $|\mathcal{B}|$ \hfill$\triangleright$ Increment number of seen samples with minibatch size
        \ENDFOR
      \ENDFOR
    \ENDFOR
    \STATE \textbf{return:} updated memory buffer {$\mathcal{M}$}
\end{algorithmic}
\end{small}
\end{algorithm}

\begin{figure*}[tb]
\begin{center}
\begin{tabular}{ccc}
\includegraphics[width=0.3155\textwidth]{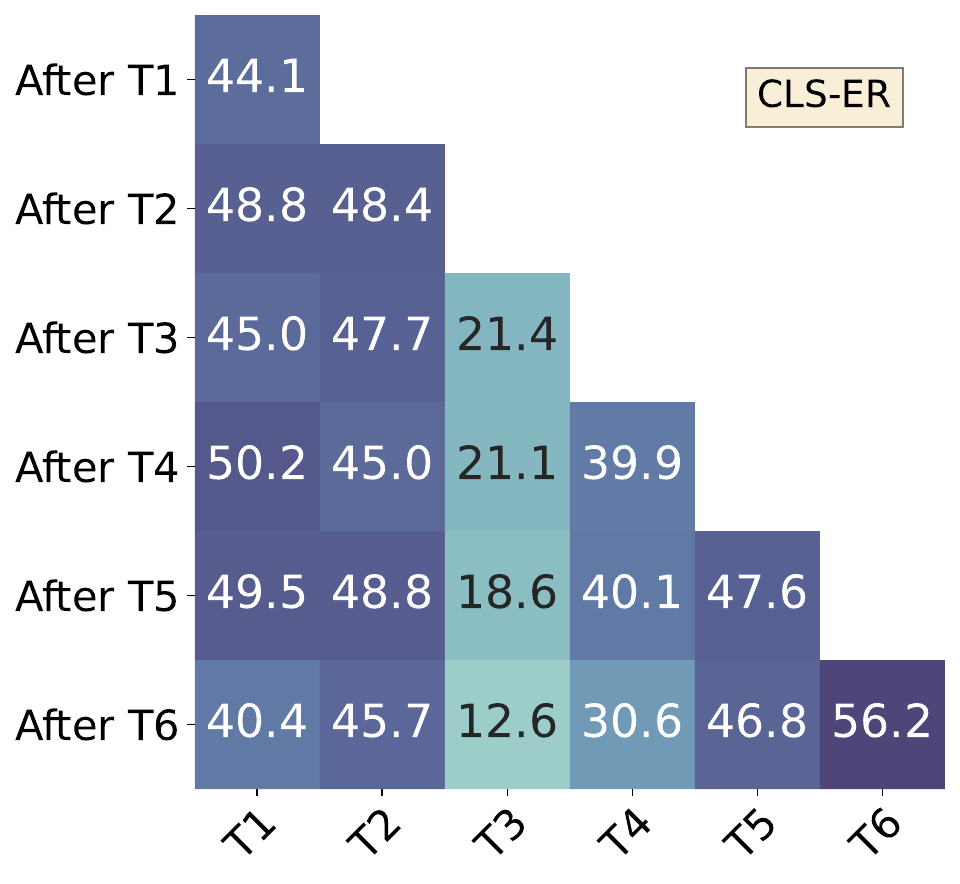} & 
\includegraphics[width=0.29\textwidth]{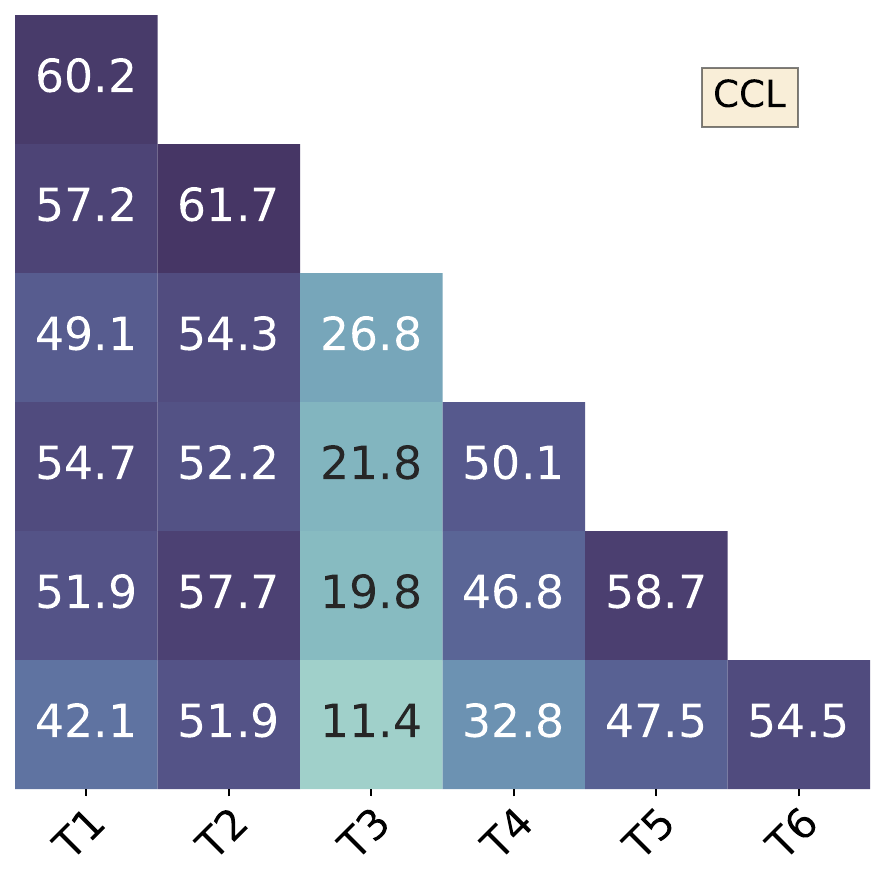} &
\includegraphics[width=0.29\textwidth]{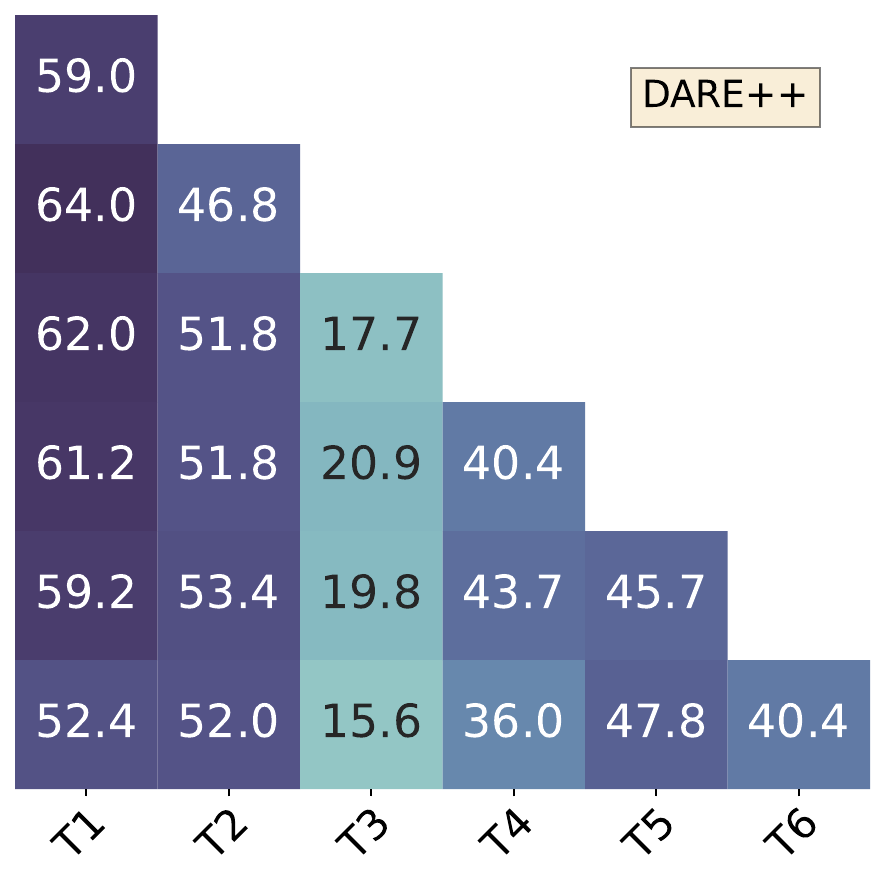}
\end{tabular}
\caption{Task-wise accuracy of different multi-memory CL models while learning new tasks with buffer size 50 on DN4IL. DARE++ performs on par with the Stable Model of CLS-ER and the Semantic Memory of DUCA, which require more training memory and computations.}
\label{fig:task_multimodel}
\end{center}
\end{figure*}



We propose the ``\textit{Intermediary Reservoir Sampling (IRS)}" strategy, which employs a normal distribution on the learning trajectory of each task. The mean of the distribution is set to the intermediate stages, and the buffer is populated accordingly. This incentivizes the storage of logits with more "dark knowledge" about the current task, which in turn propagates the knowledge across future tasks through distillation. Algorithm \ref{alg:dark_logit} describes the IRS strategy. 

Furthermore, we probe the improvements brought about by the proposed approach over DARE and DER++ in Table \ref{tab:sampling_ablation}. It is evident that the proposed IRS strategy improves both DER++ and DARE in the most challenging learning setting with a buffer size of 50. IRS improves DER++ by $\sim$5\% and DARE by $\sim$12\%.

\begin{table*}[t]
\centering
\caption{Results on DIL benchmarks with varying buffer sizes averaged over three runs. DARE achieves a consistent improvement over the other methods across different metrics, i.e., accuracy and BWT. Accuracy determines the performance on all tasks learned by the model, and backward transfer quantifies the degree to which learning a new task improves performance on previously learned tasks.}
\label{tab:extended}
\begin{small}
\begin{tabular}{cl|ccc|ccc}
\hline
\multirow{2}{*}{Buffer Size} & \multirow{2}{*}{Method} & \multicolumn{3}{c|}{iCIFAR-20} & \multicolumn{3}{c}{DN4IL} \\ \cline{3-8}
 &  & \#P $\downarrow$ & BWT $\uparrow$ & Last Accuracy $\uparrow$ & \#P $\downarrow$ & BWT $\uparrow$ & Last Accuracy $\uparrow$ \\
\hline
\multirow{4}{*}{-} & Joint & 11.18 & - & 79.61\tiny{$\pm$0.13} & 11.22 & -  & 59.93\tiny{$\pm$1.07} \\
 & SGD & 11.18 & -43.72\tiny{$\pm$1.07} & 49.40\tiny{$\pm$0.53} & 11.22 & -42.42\tiny{$\pm$0.00}  & 21.63\tiny{$\pm$0.42} \\
 \cline{2-8}
 & oEWC & 11.18 & -41.35\tiny{$\pm$2.01} & 47.39\tiny{$\pm$2.00} & 11.22 & -38.42\tiny{$\pm$0.57}  & 19.56\tiny{$\pm$1.05} \\
 & SI & 11.18 & -41.44\tiny{$\pm$2.75} & 45.94\tiny{$\pm$2.48} & 11.22 & -25.20\tiny{$\pm$2.75}  & 21.67\tiny{$\pm$1.47} \\
 \hline
 \multirow{5}{*}{50} & ER & 11.18 & -42.03\tiny{$\pm$0.27} & 50.23\tiny{$\pm$0.94} & 11.22  & -36.11\tiny{$\pm$0.26} & 24.24\tiny{$\pm$0.34} \\
 & A-GEM & 11.18 & -43.02\tiny{$\pm$0.88} & 50.02\tiny{$\pm$0.14} & 11.22 & -35.38\tiny{$\pm$0.35} & 27.06\tiny{$\pm$0.35} \\
 & FDR & 11.18 & -42.05\tiny{$\pm$1.57} & 51.07\tiny{$\pm$0.58} & 11.22 & -38.48\tiny{$\pm$1.02} & 25.09\tiny{$\pm$0.66} \\
 & DER++ & 11.18 & -40.63\tiny{$\pm$0.49} & 52.68\tiny{$\pm$1.10} & 11.22 & -29.05\tiny{$\pm$1.35} & 28.08\tiny{$\pm$0.99} \\
 & DARE & 11.19 & \textbf{-35.64}\tiny{$\pm$0.00} & \textbf{53.18}\tiny{$\pm$1.00} & 11.27 &  \textbf{-22.98}\tiny{$\pm$0.62} & \textbf{32.32}\tiny{$\pm$0.53} \\
 \hline
 \multirow{5}{*}{100} & ER & 11.18 & -41.88\tiny{$\pm$0.59} & 50.85\tiny{$\pm$0.73} &  11.22 & -35.28\tiny{$\pm$1.20} & 24.67\tiny{$\pm$0.86} \\
 & A-GEM & 11.18 & -42.98\tiny{$\pm$0.80} & 50.43\tiny{$\pm$0.57} & 11.22 & -35.78\tiny{$\pm$0.08} & 27.15\tiny{$\pm$0.33} \\
 & FDR & 11.18 & -41.22\tiny{$\pm$0.58} & 52.37\tiny{$\pm$0.39} & 11.22 & -37.26\tiny{$\pm$0.56} & 26.08\tiny{$\pm$0.65} \\
 & DER++ & 11.18 & -37.33\tiny{$\pm$1.47} & 55.32\tiny{$\pm$0.69} & 11.22 & -27.78\tiny{$\pm$0.90} & 32.06\tiny{$\pm$1.05} \\
 & DARE & 11.19 & \textbf{-33.20}\tiny{$\pm$0.09} & \textbf{56.01}\tiny{$\pm$0.22} & 11.27 & \textbf{-19.37}\tiny{$\pm$0.43} & \textbf{37.16}\tiny{$\pm$0.62} \\
 \hline
\multirow{5}{*}{200} & ER & 11.18 & -38.98\tiny{$\pm$0.74} & 52.57\tiny{$\pm$0.79} & 11.22  & -32.35\tiny{$\pm$0.51} & 27.45\tiny{$\pm$0.94} \\
 & A-GEM & 11.18 & -41.49\tiny{$\pm$0.75} & 51.12\tiny{$\pm$0.76} & 11.22 & -35.65\tiny{$\pm$0.05} & 27.44\tiny{$\pm$0.39}  \\
 & FDR & 11.18 & -38.82\tiny{$\pm$0.85} & 54.06\tiny{$\pm$0.61} & 11.22 & -36.26\tiny{$\pm$0.55} & 27.21\tiny{$\pm$0.53}\\
 & DER++ & 11.18 & -33.61\tiny{$\pm$0.64} & 58.39\tiny{$\pm$0.38} & 11.22 & -23.99\tiny{$\pm$0.74} & 35.74\tiny{$\pm$0.67} \\
 & DARE & 11.19 & \textbf{-30.22}\tiny{$\pm$1.84} & \textbf{58.53}\tiny{$\pm$1.25} & 11.27 & \textbf{-14.69}\tiny{$\pm$0.19}  & \textbf{40.59}\tiny{$\pm$0.73} \\
\hline
\end{tabular}
\end{small}
\end{table*}

\subsection{Task-wise Performance for Multi-Memory Methods}

We analyzed the task-wise accuracies of the single-model versions in the main text, and here we compare DARE++ with other multi-memory methods like CLS-ER and DUCA. 
Figure \ref{fig:task_multimodel} shows the task-wise accuracy of different CL approaches while learning new tasks. It can be seen that DARE++ retains accuracy on the initial tasks much better than CLS-ER and DUCA.
DARE++ effectively consolidates the knowledge about the past tasks from the working model compared to other memory-intensive approaches, and this can be attributed to DARE inhibiting the excessive representation of past tasks, and thus retaining the performance on them.
It should be noted that DARE++ performs well despite requiring less memory and training computations, as evident from the number of parameters in Table 1. This makes it more efficient and effective in real-time applications.

\begin{table*}[t]
\centering
\caption{Selected hyperparameters for DARE and DARE++.}
\label{tab:hparams_dare}
\begin{small}
\begin{tabular}{llc|cccccccc}
\hline
 Dataset & Method & Buffer Size & lr & $\alpha$ & $\beta$ & sw & st & r & $\alpha$ \\
 \hline
\multirow{6}{*}{iCIFAR-20} & \multirow{3}{*}{DARE} & 50 & 0.04 & 0.3 & 0.1 & 0.05 & 1.2 & - & - \\
 &  & 100 & 0.03 & 0.5 & 0.1 & 0.08 & 1.0 & - & - \\
 &  & 200 & 0.06 & 0.5 & 0.1 & 0.08 & 0.99 & - & - \\
 \cline{2-10}
 & \multirow{3}{*}{DARE++} & 50 & 0.04 & 0.3 & 0.1 & 0.05 & 1.2 & 0.055 & 0.999 \\
 &  & 100 & 0.03 & 0.5 & 0.1 & 0.08 & 1.1 & 0.058 & 0.999 \\
 &  & 200 & 0.06 & 0.5 & 0.2 & 0.08 & 1.1 & 0.045 & 0.999 \\
 \hline
\multirow{6}{*}{DN4IL} & \multirow{3}{*}{DARE} & 50 & 0.04 & 0.1 & 0.2 & 0.05 & 0.8 & - & - \\
 &  & 100 & 0.04 & 0.1 & 1.0 & 0.05 & 0.8 & - & - \\
 &  & 200 & 0.04 & 0.1 & 1.0 & 0.05 & 0.8 & - & - \\
\cline{2-10}
 & \multirow{3}{*}{DARE++} & 50 & 0.04 & 0.1 & 0.2 & 0.05 & 0.8 & 0.050 & 0.999 \\
 &  & 100 & 0.04 & 0.1 & 1.0 & 0.05 & 0.8 & 0.050 & 0.999 \\
 &  & 200 & 0.04 & 0.1 & 1.0 & 0.05 & 0.8 & 0.090 & 0.999 \\
 \hline
\end{tabular}
\end{small}
\end{table*}

\subsection{Extended Results with Conventional CL Methods}

In addition to the comparisons in the main text, we compare DARE with conventional regularization- and replay-based methods. Online EWC~\cite{schwarz2018progress} and Synaptic Intelligence~\cite{zenke2017continual} fall under regularization-based methods, where changes to important parameters in the network for old tasks are penalized. Averaged-Gradient Episodic Memory (A-GEM)~\cite{https://doi.org/10.48550/arxiv.2011.07801} and Function Distance Regularization (FDR)~\cite{benjamin2018measuring} fall into replay-based methods. A-GEM learns new tasks with an optimization constraint such that the gradients for new tasks are projected to the orthogonal subspace of the gradients for old task samples, thus retaining the performance on old tasks. FDR applies distillation loss with respect to network outputs stored in the buffer for past task samples.

Table \ref{tab:extended} compares DARE with various baselines on different datasets and buffer sizes. Regularization-based methods (oEWC and SI) face challenges in the DIL scenario, while A-GEM and FDR demonstrate performance comparable to ER. It is evident that DARE surpasses all other baselines in all settings.

\begin{table*}[t]
\centering
\caption{Selected hyperparameters for all baselines.}
\label{tab:hparams}
\begin{small}
\begin{tabular}{lcl|l}
\hline
Dataset & Buffer Size & Method & \multicolumn{1}{c}{Hyperparameters} \\
\hline
\multirow{20}{*}{iCIFAR-20} & \multirow{2}{*}{-} & oEWC & \emph{lr=}0.03, $\lambda$=10, $\gamma$=1 \\
 &  & SI & \emph{lr=}0.03, $c$=1, $\xi$=0.9 \\
 \cline{2-4}
 & \multirow{6}{*}{50} & ER & \emph{lr=}0.1 \\
 &  & A-GEM & \emph{lr=}0.05 \\
 &  & FDR & \emph{lr=}0.03, $\alpha$=0.1 \\
 &  & DER++ & \emph{lr=}0.03, $\alpha$=0.1, $\beta$=0.2 \\
 &  & CLS-ER & \emph{lr=}0.05, $\lambda$=0.1, $r_p$=0.06, $r_s$=0.02, $d_p$=0.999, $d_s$=0.999 \\
 &  & DUCA & \emph{lr=}0.05, $\lambda$=0.1, $r$=0.08, $d$=0.999 \\
 \cline{2-4}
 & \multirow{6}{*}{100} & ER & \emph{lr=}0.1 \\
 &  & A-GEM & \emph{lr=}0.06 \\
 &  & FDR & \emph{lr=}0.03, $\alpha$=0.2 \\
 &  & DER++ & \emph{lr=}0.05, $\alpha$=0.1, $\beta$=0.1 \\
 &  & CLS-ER & \emph{lr=}0.03, $\lambda$=0.1, $r_p$=0.08, $r_s$=0.04, $d_p$=0.999, $d_s$=0.999 \\
 &  & DUCA & \emph{lr=}0.04, $\lambda$=0.1, $r$=0.09, $d$=0.999 \\
 \cline{2-4}
 & \multirow{6}{*}{200} & ER & \emph{lr=}0.1 \\
 &  & A-GEM & \emph{lr=}0.04 \\
 &  & FDR & \emph{lr=}0.03, $\alpha$=0.5 \\
 &  & DER++ & \emph{lr=}0.03, $\alpha$=0.2, $\beta$=0.1 \\
 &  & CLS-ER & \emph{lr=}0.05, $\lambda$=0.1, $r_p$=0.12, $r_s$=0.04, $d_p$=0.999, $d_s$=0.999 \\
 &  & DUCA & \emph{lr=}0.04, $\lambda$=0.1, $r$=0.08, $d$=0.999 \\
 \hline
 \multirow{20}{*}{DN4IL} & \multirow{2}{*}{-} & oEWC & \emph{lr=}0.05, $\lambda$=50, $\gamma$=1 \\
 &  & SI & \emph{lr=}0.05, $c$=0.5, $\xi$=0.5 \\
 \cline{2-4}
 & \multirow{6}{*}{50} & ER & \emph{lr=}0.1 \\
 &  & A-GEM & \emph{lr=}0.05 \\
 &  & FDR & \emph{lr=}0.03, $\alpha$=0.5 \\
 &  & DER++ & \emph{lr=}0.01, $\alpha$=0.1, $\beta$=0.1 \\
 &  & CLS-ER & \emph{lr=}0.05, $\lambda$=0.1, $r_p$=0.06, $r_s$=0.04, $d_p$=0.999, $d_s$=0.999 \\
 &  & DUCA & \emph{lr=}0.04, $\lambda$=0.1, $r$=0.06, $d$=0.999 \\
 \cline{2-4}
 & \multirow{6}{*}{100} & ER & \emph{lr=}0.1 \\
 &  & A-GEM & \emph{lr=}0.04 \\
 &  & FDR & \emph{lr=}0.05, $\alpha$=0.5 \\
 &  & DER++ & \emph{lr=}0.03, $\alpha$=0.2, $\beta$=0.5 \\
 &  & CLS-ER & \emph{lr=}0.05, $\lambda$=0.1, $r_p$=0.14, $r_s$=0.04, $d_p$=0.999, $d_s$=0.999 \\
 &  & DUCA & \emph{lr=}0.05, $\lambda$=0.1, $r$=0.06, $d$=0.999 \\
 \cline{2-4}
 & \multirow{6}{*}{200} & ER & \emph{lr=}0.1 \\
 &  & A-GEM & \emph{lr=}0.04 \\
 &  & FDR & \emph{lr=}0.05, $\alpha$=0.1 \\
 &  & DER++ & \emph{lr=}0.03, $\alpha$=0.1, $\beta$=1.0 \\
 &  & CLS-ER & \emph{lr=}0.05, $\lambda$=0.1, $r_p$=0.08, $r_s$=0.04, $d_p$=0.999, $d_s$=0.999 \\
 &  & DUCA & \emph{lr=}0.03, $\lambda$=0.1, $r$=0.06, $d$=0.999 \\
 \hline
\end{tabular}
\end{small}
\end{table*}

\subsection{Hyperparameters}

We enumerate the best hyperparameters chosen for the evaluation of different methods in the main paper and Table \ref{tab:extended}. $lr$ denotes the learning rate for the entire learning trajectory in each task. We fixed the batch size to 32 for both the current task and old task samples (in buffer memory). We used grid search to find the best hyperparameters and took reference from \emph{mammoth}~\cite{buzzega2020dark} repository for the search for experiments on iCIFAR-20 dataset and DUCA~\cite{gowda2023a} for the search for experiments on DN4IL dataset, respectively. We trained all methods using SGD optimizer for 50 epochs per task.

Table \ref{tab:hparams_dare} outlines the hyperparameters chosen for DARE and DARE++, while Table \ref{tab:hparams} lists the hyperparameters selected for various CL baselines in our study. $r$ denotes the frequency of updating the semantic model from the working model, and $d$ denotes the rate at which the weights of the EMA model are updated. $r_p$ and $r_s$ stand for the update frequency for the plastic and stable model, respectively, in CLS-ER. $\lambda$ refers to the weighting parameter for the knowledge distillation from the semantic model to the working model in DUCA and CLS-ER. Furthermore, $sw$ and $st$ denote the weight and temperature used in the supervised contrastive loss.

The losses corresponding to \emph{Divergence} and \emph{Adaptation} steps in DARE/DARE++ were weighted by 0.1 and 1, respectively, in all datasets and buffer sizes. It is also evident from Table \ref{tab:hparams_dare} that DARE and DARE++ do not need extensive finetuning, except for the dataset-specific learning rates. The other hyperparameters are mostly stable across different settings.

\end{document}